\documentclass[conference]{IEEEtran}
\IEEEoverridecommandlockouts
\usepackage{cite}
\usepackage{amsmath,amssymb,amsfonts}
\usepackage{algorithmic}
\usepackage{graphicx}
\usepackage{textcomp}
\usepackage{xcolor}
\usepackage{booktabs}
\usepackage{times}  
\usepackage{helvet} 
\usepackage{courier}  
\usepackage{algorithm}
\usepackage{algorithmic}
\usepackage{amsmath}
\usepackage{amsthm}
\usepackage{amssymb}
\usepackage{booktabs}
\usepackage{bbding}
\usepackage{xspace} 
\usepackage{rotating}
\usepackage{multirow}
\usepackage{array}   
\usepackage{rotating} 
\usepackage{colortbl} 
\usepackage{arydshln} 
\usepackage{epigraph}
\usepackage[capitalize]{cleveref} 
\crefname{section}{Sec.}{Secs.}
\Crefname{section}{Section}{Sections}
\Crefname{table}{Table}{Tables}
\crefname{table}{Tab.}{Tabs.}

\newcommand{\eg}{\emph{e.g.,}\xspace}
\newcommand{\wrt}{\emph{w.r.t.}\xspace}
\newcommand{\ie}{\emph{i.e.,}\xspace}
\newcommand{\etc}{\emph{etc.}\xspace}

\def\BibTeX{{\rm B\kern-.05em{\sc i\kern-.025em b}\kern-.08em
    T\kern-.1667em\lower.7ex\hbox{E}\kern-.125emX}}

\begin{document}

\title{Overcoming Language Priors for Visual Question Answering Based on Knowledge Distillation\\
}

\author{
\IEEEauthorblockN{Daowan Peng\textsuperscript{\rm 1,\rm 2}, Wei Wei\textsuperscript{\rm 1,\rm 2}\thanks{$^{\ast}$Corresponding author.}$^{\ast}$}
\IEEEauthorblockA{\textsuperscript{\rm 1}\textit{CCIIP Lab, School of Computer Science and Technology, Huazhong University of Science and Technology}}
\IEEEauthorblockA{\textsuperscript{\rm 2}\textit{Joint Laboratory of HUST and Pingan Property \& Casualty Research (HPL)}}
\IEEEauthorblockA{Wuhan, China}
\IEEEauthorblockA{pengdw@hust.edu.cn, weiw@hust.edu.cn}
}

\maketitle

\begin{abstract}

Previous studies have pointed out that visual question answering (VQA) models are prone to relying on language priors for answer predictions. In this context, predictions often depend on linguistic shortcuts rather than a comprehensive grasp of multimodal knowledge, which diminishes their generalization ability. 
In this paper, we propose a novel method, namely, KDAR, leveraging knowledge distillation to address the prior-dependency dilemmas within the VQA task. 
Specifically, the regularization effect facilitated by soft labels from a well-trained teacher is employed to penalize overfitting to the most common answers. The soft labels, which serve a regularization role, also provide semantic guidance that narrows the range of candidate answers. Additionally, we design an adaptive sample-wise reweighting learning strategy to further mitigate bias by dynamically adjusting the importance of each sample. 
Experimental results demonstrate that our method enhances performance in both OOD and IID settings. Our method achieves state-of-the-art performance on the VQA-CPv2 out-of-distribution (OOD) benchmark, significantly outperforming previous state-of-the-art approaches.

\end{abstract}


\section{Introduction}
\label{sec:intro}

\begin{figure}[!t]
	\centering 
	\includegraphics[width = .46\textwidth,height=0.2\textwidth]{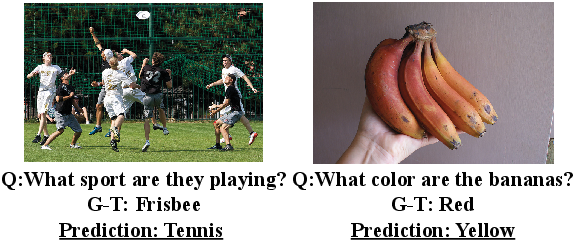}
	\caption{The previous VQA models tend to predict the most frequent answers in training dataset based on prior knowledge of question-answer pairs, rather a fine-grained understanding of both images and questions.}
	\label{Fig1}
\end{figure}

As a high-level task that bridges the gap between computer vision and natural language processing, visual question answering (VQA) \cite{antol2015vqa, anderson2018bottomup,liu2022depth,gong2022vqamixct} has received critical attention from artificial intelligence communities. With the advancements in multi-modal machine learning, the VQA tasks have yielded ongoing success. However, recent studies \cite{han2023general, agrawal2018dont} reveal that the VQA models tend to exploit the spurious statistical correlations between the questions and answers, rather than genuinely understanding questions based on image content. 
For instance, as illustrated in \cref{Fig1}, when presented with the question \textit{``What color are the bananas?''}, most advanced models predict the common answer \textit{`yellow'}, irrespective of the image content. Since most bananas are \textit{`yellow'} in the training dataset. Similarly, when answering a question like \textit{``What sport ...?''}, VQA models usually predict the most frequent answer \textit{`tennis'}.
Unfortunately, models remember such spurious correlations (\ie language prior or language bias) would decrease their generalization ability in out-of-distribution (OOD) scenarios.

Nowadays, numerous efforts have been made to address the language bias issue from different perspectives \cite{cadene2019rubi,chen2020counterfactual,zhu2020overcoming,goyal2017making,han2021greedy,wen2021debiased}. 
For instance, \cite{goyal2017making} balance training samples by collecting more similar images that would force VQA models to focus on visual information.  \cite{selvaraju2019taking} and \cite{wu2019self-critical} exploit additional annotations to make the VQA models look at relevant visual regions, instead of making use of linguistic priors before producing answers. \cite{liang2020learning,chen2020counterfactual,chen2023ccst}  suggest that the generated counterfactual samples can help the model to learn well.  
Another popular line of solutions focuses on designing the structure of the model to encourage a better use of image and language information to diminish existing idiosyncratic biases in datasets. For example, \cite{cadene2019rubi} and \cite{clark2019dont} employ a question-only branch to capture the language bias in the train stage, while excluding the captured bias during the inference period. 
However, \cite{wen2021debiased} suggest that removing all the language priors may harm the performance since not all the biases are detrimental to the model.
GGE \cite{han2021greedy} and GGD \cite{han2023general} have identified that the language bias in VQA merges from two main aspects, namely, \textit{distribution bias} and  \textit{shortcut bias} respectively. The \textit{distribution bias} is in terms of the answer distributions of different question types which are mainly presented as long-tailed. While the \textit{shortcut bias} is caused by the strong correlation between the questions and the answers. 
As a result, the models usually produce right answers for the wrong reasons, \ie VQA models give an answer based on the question-answer pairs rather than focusing on the correct  visual regions. 

Building upon the research findings of GGE and GGD, we delve deeper into the language prior issue in the VQA task, approaching it from a novel perspective. We conceptualize the language prior problem as a fitting problem in machine learning, wherein the model tends to overfit the most frequently occurring samples while underfitting  rare samples.
Specifically, this tendency is evident that the VQA models are inclined to overly learn --- and often overfit --- the most common samples \wrt the question types in the training dataset while paying less attention to rare samples. 
As a result,  VQA models tend to produce the most common answers with high confidence due to the high co-occurrence factor in the training dataset (\eg \textit{What color ... bananas} $\to$ \textit{yellow}; \textit{What sport ...}  $\to$ \textit{tennis}, \etc). 

To address this issue, we propose two learning strategies. 
Firstly, we introduce a common machine learning strategy used to combat overfitting –-- regularization. Specifically, we incorporate soft labels from knowledge distillation to help alleviate the overfitting problem of VQA models on common samples, with soft labels acting as a regularizer. Additionally, soft labels provide extra supervisory signals with inter-class similarity, assisting the VQA model in filtering out irrelevant answers, narrowing down the range of candidate answers, and thereby improving model accuracy. Secondly, to enhance the model's focus on rare samples while further reducing its emphasis on common samples, we propose a novel, adaptive sample-wise reweighting learning strategy. This strategy dynamically adjusts the loss weights for each sample based on the model's learning progress. 
Based on the two learning strategies, we name our approach KDAR.
To sum up, our main contributions are as the following three-fold:
\begin{itemize}

\item We approach the language prior issue in the VQA task from a new perspective, conceptualizing it as a fitting problem in machine learning.

\item We introduce KDAR, a simple yet effective method, to mitigate the impact of language priors. KDAR uses soft labels as a regularizer to tackle overfitting with common samples and dynamically adjusts the loss weights for each sample, promoting a balanced learning process that can prevent the most common answers from dominating the correct answers.

\item  We evaluate the proposed method on the VQA-CPv2 and VQAv2 datasets, and the experimental results show that our method outperforms the previous methods by a large margin. Especially, when combined with LXMERT, our method achieves the best overall accuracy of 71.33\% on VQA-CPv2 dataset. 
\end{itemize}

\section{Related Works}
As a multi-modal task, VQA needs a full understanding of both images and questions \cite{liu2022declaration,liu2024detection}. Nevertheless, a variety of state-of-the-art models are inclined to leverage the language priors shortcuts, which makes the VQA task notorious due to the harmful effects on the generalization to OOD datasets. Fortunately, research scholars have proposed a variety of approaches to deal with this issue. These approaches can be broadly divided into two camps: augmentation-based methods and non-augmentation-based methods.

\subsection{Augmentation-based methods}
To encourage VQA models to gear towards the image grounding rather than spurious correlations, HINT \cite{selvaraju2019taking} and SCR \cite{wu2019self-critical} apply additional annotations. HINT exploits human-based attention maps that force models to base their decisions on these same regions with human-based ones. While SCR utilizes human visual/textual explanations to influential regions, moreover, it criticizes the sensitivity of the wrong answer to the important regions. Consequently, the model could be more sensitive to relevant visual regions. 
\cite{chen2020counterfactual, chen2023ccst} generates counterfactual samples by masking the critical object in the image or the keyword in question, hence the models focus on all important objects in image and keywords in question which is beneficial to alleviate the language priors.
SSL \cite{zhu2020overcoming} decreases the language priors in a self-supervised fashion, which replaces the image of the question-answer-image triple tuple randomly and maximizes the loss of the generated irrelevant pairs. 
\cite{kil2021discovering} proposes a data augmentation pipeline to convert implicit knowledge in the dataset into explicit training examples.
\cite{chen2022rethinking} argues the mainstream data augmentation strategies are synthetic-based methods by editing some visual regions or words, which are unnatural and error-prone. They proposed the KDDAug based on knowledge distillation to generate pseudo-answers.
\cite{huai2023sqt} constructs negative samples by shuffling the question types of the original questions. 
\cite{peng2023empirical} proposes treating the questions of shuffled sequences as data augmentation for modeling purposes.
\cite{wen2023digging} constructs both positive and negative samples in vision and language modalities.

\subsection{Non-augmentation-based methods}
Unlike the aforementioned augmentation-based ones, non-augmentation-based methods mainly focus on design strategy based on the model itself.
RUBi \cite{cadene2019rubi} and LMH \cite{clark2019dont} re-weight samples based on a question-only branch, combined with a classic VQA model, to capture the language bias during the training process. 
\cite{guo2022loss} proposes a loss rescaling approach to reweight the samples during the training stage.
DLR \cite{jing2020overcoming} overcomes language bias by decomposing the question into different regions using a language attention module and leveraging the corresponding representations to infer answers. By introducing the cause-effect,  the CF-VQA \cite{niu2021counterfactual}  treats the language bias as the direct causal effect of questions on answers and subtracts the direct language effect during the inference period.  By using a divide-and-conquer strategy, GGE \cite{han2021greedy} fits the biased feature with multiple biased models and uses a base model for the remaining unbiased feature. 
\cite{guo2021adavqa, basu2023rmlvqa} de-biases the language priors from the viewpoint of the feature space learning via manipulating the learned feature space of answers with adaptive margin loss.
D-VQA \cite{wen2021debiased} alleviates the ``negative'' biases of both visual and language modalities based on sample perspective and feature perspective. 
IntroD \cite{niu2021introspective} improves the performance by capturing the in-distribution and out-of-distribution inductive bias.
\cite{cho2023generative} utilizes a generative network to mimic biases in the target model by combining adversarial objectives and knowledge distillation.
In this work, we leverage the soft labels of the knowledge distillation scheme to de-biase language priors in the VQA task.

\section{Methodology}

\subsection{Preliminaries}
\subsubsection{The Paradigm of VQA}
The VQA task is a multi-modal task regarding visual modal and language modal, given an image $v \in{\mathcal{I}} $,  and a question $q \in{\mathcal{Q}} $ about the image, VQA model aims to predict an answer $a\in{\mathcal{A}}$ that can best match the ground-truth answer \textit{a*}. This can be defined as a classification task, without loss of generality, a generic VQA model can be formulated as
\begin{equation}
a = f(v,q)=h(g(e_v(v),e_q(q))),
\label{eq:vqa}
\end{equation}   
where $e_v(\cdot)$ and $e_q(\cdot)$ denote image encoder and question encoder, respectively. $g(\cdot)$ is a multi-modal feature fusion function, and $h(\cdot)$ is a multi-layer perception classifier. The VQA task tries to minimize the following binary cross-entropy (BCE) loss function:
\begin{equation}
\mathcal{L}_{bce} = -\frac{1}{N} \sum_{i}^{N}\big[y_{i}log(p_i)+(1-y_i)log(1-p_i)\big],
\label{eq:loss_vqa}
\end{equation}  
where $p_i$ is the prediction, $y_{i}$ is the ground-truth label, and $N$ denotes the number of samples in the dataset. 

\subsubsection{The Paradigm of Knowledge Distillation}
We briefly introduce knowledge distillation, which aims to transfer knowledge from a teacher network to a student network. 
The process of knowledge transfer is achieved through the student model, which mimics the output probability distribution of the teacher model.
The distillation loss of knowledge distillation framework \cite{hinton2015distilling} can be formulated as,
\begin{gather}
\begin{aligned}
    \mathcal{L} = -\tau^2 D_{KL} (p^t_\tau, p^s_\tau) \label{eq:ori_kd} \\
\end{aligned}
\end{gather}
where $p^t_\tau$ and $p^s_\tau$ denote the outputs of the teacher model and student model, respectively. $D_{KL}$ is the Kullback-Leible (KL) divergence, and $ \tau $ is the temperature hyperparameter.

\subsection{KDAR Method}
\begin{figure}[t]
	\centering
	\includegraphics[width =.47\textwidth,height=0.2\textwidth]{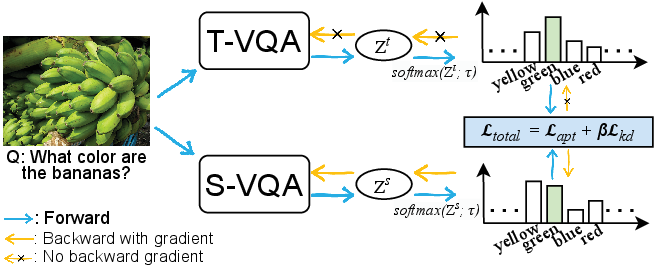}
	\caption{The training pipeline of our method. The image-question pairs are fed to a well-trained teacher VQA (T-VQA) model and the student VQA (S-VQA) model. Then the softmax function is used to the logits $Z^t$ of the teacher model and $Z^s$ of the student model to compute the probabilities at a temperature $\tau$. Thereafter, the student model needs to minimize the total loss. Note that the parameters of the T-VQA will be frozen during training stage.}  
	\label{Fig2}
\end{figure}

In this section, we will go into detail on our proposed KDAR approach. Firstly, we explain the whys and wherefores of the \underline{K}nowledge \underline{D}istillation learning strategy and then we illustrate how the soft labels of the teacher model work. After that, we introduce the \underline{A}daptive \underline{R}eweighting learning strategy. The training pipeline is shown in \cref{Fig2}.

\subsubsection{Knowledge Distillation Learning Strategy}

In the paradigm of knowledge distillation, the output probability distribution $p^t_\tau$ from the teacher model is referred to as soft labels.  In contrast to conventional one-hot label $y$, soft labels offer a richer source of supervisory signals, which serve as a form of label smoothing \cite{yuan2020revisiting,shen2020label}. Specifically, label smoothing in knowledge distillation can be regarded as a replacement of the one-hot label $y$, with a mixture of $y$ and the output probability distribution $p^t_\tau$ of the teacher model in the form of $y' = (1-\alpha) y + \alpha p^t_\tau$, where the $\alpha$ is a smoothing factor. The cross-entropy loss $H(y', p^s_\tau)$ defined over the mixture $y'$ is 
\begin{equation}
\begin{aligned}
H(y', p^s_\tau) &= H\big((1-\alpha) y + \alpha p^t_\tau, p^s_\tau\big)\\
&=(1-\alpha) H(y, p^s_\tau)+\alpha H(p^t_\tau, p^s_\tau)\\
& = (1-\alpha) H(y, p^s_\tau)+\alpha(D_{KL}(p^t_\tau, p^s_\tau) + H(p^t_\tau)).
\end{aligned}
\end{equation}   
where $D_{KL}$ measures the KL divergence between the output probability $p^t_\tau$ of the teacher model and the output probability $p^s_\tau$ of the student model, $H(p^t_\tau)$ denotes the entropy of $p^t_\tau$ and is a constant.  Therefore, the loss function can be re-written as 
\begin{equation}
\mathcal{L}_{kd} = H(y', p^s_\tau)= (1-\alpha) H(y, p^s_\tau)+\alpha(D_{KL}(p^t_\tau, p^s_\tau)).
\label{eq:kd}
\end{equation} 
This equation reveals that knowledge distillation essentially involves the student model simultaneously minimizing the cross-entropy with respect to the ground-truth labels and minimizing the KL divergence with respect to the teacher model. Besides, the second term in this equation,  \ie  $\alpha(D_{KL}(p^t_\tau, p^s_\tau))$ , can be regarded as a regularization term aimed at harmonizing the parameters of the student model to prevent overfitting. Therefore, the soft labels in knowledge distillation play the role of a regularizer.
Back to the VQA task, the distribution of the dataset is long-tail \cite{agrawal2018dont}, for example, yellow-colored bananas constitute the vast majority. Consequently, when the model faced with a question like \textit{``What color are the bananas?''}, the corresponding ground-truth label is usually \textit{`yellow'}. During the learning process, the model tends to best learn --- often overfit --- these head samples, while neglecting those tail samples (\eg green-colored bananas). As a result, the model often predicts \textit{`yellow'} with high confidence as the answer when faced with the question \textit{``What color ... bananas?''}. However, these traits hamper the model from acquiring more generalized knowledge during training. Therefore, we leverage the regularization effect of soft labels in knowledge distillation to alleviate overfitting issues in the VQA task and reduce the model's overemphasis on head samples.

Aside from the regularization effect, the soft labels in knowledge distillation paradigm can also provide additional supervisory signals with semantically similar meanings, further aiding the model in learning. For example, when answering the question \textit{``What color ...?''}, color-related supervisory signals (`red', `blue', `yellow', \etc) from the teacher model can help narrow down the range of candidate answers, informing the model that this question is with respect to color and preventing it from predicting non-color-related answers.

\subsubsection{Adaptive Reweighting Learning Strategy}
Given the long-tail distribution evident in the VQA dataset \cite{agrawal2018dont}, the models tend to focus more on head samples, leading to insufficient learning of tail samples. To mitigate this, we propose a novel adaptive sample-level reweighting learning strategy to reduce the model's excessive learning (overfitting) on head samples, while increasing the model's attention to tail samples simultaneously.
Specifically, based on knowledge distillation, we exploit the relative correlation between the output, $p^t_\tau$, of the teacher model and the output, $p^s_\tau$, of the student model to dynamically adjust the weight of each sample for the student model. The adaptive loss function is as follows, 
\begin{equation}
\mathcal{L}_{apt} = \left(1-exp\big(\dfrac{-log(p^t_\tau)}{log(p^s_\tau)}\big)\right)\mathcal{L}_{bce},
\label{eq:apt}
\end{equation}
where the $\mathcal{L}_{bce}$ is the learning objective in VQA task (\ie \cref{eq:loss_vqa}), the $exp()$ is the exponential function. Here, the teacher model is a well-trained debiasing model. The intuition is that if the output of the student model is better than that of the teacher, it is more likely that the student model starts to overfit the common samples in train stage.  

\vspace{-0.1cm}
\subsection{Learning Objective}
At last, during the training process, the learning objective is to minimize the total loss, 
\begin{equation}
\mathcal{L}_{total} =  \mathcal{L}_{apt} + \beta \mathcal{L}_{kd},
\label{loss:total}
\end{equation}
where $\beta$ is a scalar weight hyperparameter which balances the adaptive loss and the knowledge distillation loss.

\section{Experiments}
\subsection{Experimental Setting}
\noindent{\textbf{Datasets and Baselines}}. We evaluate our method on VQAv2 dataset \cite{goyal2017making} and the out-of-distribution benchmark VQA-CPv2 \cite{agrawal2018dont}, and compare it with vanilla methods SAN \cite{yang2016stacked}, S-MRL \cite{cadene2019rubi}, UpDn \cite{anderson2018bottomup}, GVQA \cite{agrawal2018dont} and the previous state-of-the-art methods, including augmentation-based methods, HINT, SCR, CL, SSL, CSS, AEMP, CSST, KDDAug, MUTANT, SQT, D-VQA,  DDG and the non-augmentation-based methods, AReg, RUBi, LMH,  DLR, AdaVQA, CCB, LP-Focal, Re-scaling, CF-VQA, IntroD, GGE, GGD, MDDC, RMLVQA, GenB.

\begin{table}[!ht]
  
  \centering
    \caption{ Acc.(\%) results comparisons with respect to different answer types on VQA-CPv2 test split. The \underline{second best}  and the \textbf{best} are highlighted in each column. }
    \begin{tabular}{c|lcccc}
    \toprule
          & Models  & All   & Yes/No & Num.  & Other \\
    \midrule
    \multirow{3}[2]{*}{\begin{sideways}vanilla\end{sideways}} & SAN \cite{yang2016stacked}        & 26.88 & 38.35 & 11.96 & 42.98 \\
          & S-MRL \cite{cadene2019rubi}       & 38.46 & 42.85 & 12.81 & 43.2 \\
          & UpDn \cite{anderson2018bottomup}  & 39.74 & 42.27 & 11.93 & 46.05 \\
          & GVQA \cite{agrawal2018dont} &31.30 &57.99 &13.68 &22.14\\ 
    \midrule
    \multirow{9}[2]{*}{\begin{sideways}augmentation-based\end{sideways}} 
          &HINT \cite{selvaraju2019taking} &46.73 &67.27 &10.61 &45.88\\
          &SCR \cite{wu2019self-critical}  & 48.47 &70.41 & 10.42 & 47.29\\
          &CL \cite{liang2020learning}  &59.18 &86.99	 &49.89	&47.16 \\
          &SSL \cite{zhu2020overcoming}  &57.59 & 86.53 &29.87 &50.03 \\
    
          & CSS \cite{chen2020counterfactual}     & 41.16 & 43.96 & 12.78 & 47.48 \\
          & AEMP \cite{peng2023empirical} &44.95 &54.51 &14.89 &48.18\\
          & CSST \cite{chen2023ccst}    & 56.55 & 80.45 & 36.29 & 49.58 \\
          & KDDAug \cite{chen2022rethinking}   & 60.24 & 86.13 &  \underline{55.08} & 48.08 \\
          & MUTANT \cite{gokhale2020mutant} &61.72 &88.90 &49.68 &50.78\\
          & SQT \cite{huai2023sqt}  &61.76	&87.96	&53.67	&50.26\\
          & D-VQA \cite{wen2021debiased}   &  \underline{61.91} & 88.93 & 52.32 &  \underline{50.39} \\   
          & DDG \cite{wen2023digging}     & 61.14 & 88.77 & 49.33 & 49.9 \\
    \midrule   
    
     \multirow{15}[2]{*}{\begin{sideways}non-augmentation-based\end{sideways}} 
     & AReg \cite{ramakrishnan2018overcoming}    & 41.17 & 65.49 & 15.48 & 35.48 \\
          & RUBi \cite{cadene2019rubi}    & 44.23 & 67.05 & 17.48 & 39.61 \\
          & LMH \cite{clark2019dont}     & 52.73 & 72.95 & 31.9  & 47.79 \\
          & DLR \cite{jing2020overcoming} &48.87 &70.99 &18.72 &45.57\\
          & AdaVQA \cite{guo2021adavqa}   & 54.67 & 72.47 & 53.81 & 45.58 \\
          & CCB \cite{yang2021learning}             &57.99	&86.41	&45.63	&48.76\\
	     & LP-Focal \cite{lao2021language}        &58.45	 &88.34	 &34.67	 &49.32\\
          & Re-scaling \cite{guo2022loss}   & 47.09 & 68.42 & 21.71 & 42.88 \\
          & CF-VQA \cite{niu2021counterfactual}  & 55.05 & \textbf{90.61} & 21.5 & 45.61 \\
          & IntroD \cite{niu2021introspective}  & 51.31 & 71.39 & 27.13 & 47.41 \\
          
          & GGE \cite{han2021greedy}     & 57.32 & 87.04 & 27.75 & 49.59 \\        
          & GGD \cite{han2023general}     & 59.37 & 88.23 & 38.11 & 49.82 \\
          & MDDC \cite{li2023multi}    & 54.7  & 83.58 & 19.93 & 49.1 \\ 
          & RMLVQA \cite{basu2023rmlvqa}  & 60.41 & \underline{89.98} & 45.96 & 48.74 \\
          & GenB \cite{cho2023generative}   & 59.15 & 88.03 & 40.05 & 49.25 \\
    \midrule
          & \textbf{Ours}  &\textbf{62.86}  &89.18   &\textbf{55.86}    &\textbf{50.98}  \\
    \bottomrule
    \end{tabular}%
  \label{tab:vqacpv2}%
\end{table}%

\noindent{\textbf{Implementations Details}}. Our method is model-agnostic, and any other VQA model could be the teacher network. To better demonstrate the effectiveness of our KDAR, we employ a de-biased VQA model as our teacher model. In experiments, we choose the best-performing D-VQA \cite{wen2021debiased} as the teacher model. The student model is UpDn, which combined our loss term solely with the sample loss of the teacher (\ie $\mathcal{L}_{sam}$ in \cite{wen2021debiased}). During the training stage, we set $\tau=2.5$ as the temperature constant, follow the experimental settings of the teacher, and do NOT change other settings, such as the learning rate, batch size, or optimizer. The hyperparameter $\beta$ in \cref{loss:total} is set as 3 in our implementations.

\noindent{\textbf{Evaluation Metric}}. In this paper, we also follow the standard metric \cite{antol2015vqa} in VQA task for evaluation, 
\begin{equation}
Acc(a) = min\left(1,\frac{\# humans\; provided \; answers}{3}\right).
\label{eq:metric}
\end{equation}   
Note that, there are ten answers for each question which are provided by human annotators.

\subsection{Experimental Results}
\subsubsection{Performance on VQA-CPv2}
\noindent{\textbf{Comparison with state-of-the-arts}}. 
The experimental results are depicted in \cref{tab:vqacpv2}, we have the following observations, (1) our method achieves the best overall accuracy over all the compared methods, including the augmentation-based methods and non-augmentation-based methods, (2) our method performs best when answering the more challenging questions, \ie on the ``Num.'' and ``Other'' questions, which indicates that our method can truly boost the performance by debiasing the language bais.

\noindent{\textbf{Effect of different backbones.}}
To further demonstrate the scalability of our approach, we conduct additional experiments based on different backbones including SAN \cite{yang2016stacked}, UpDn \cite{anderson2018bottomup} and LXMERT \cite{tan2019lxmert}. The LXMERT is a multimodal pre-trained model that has shown powerful performances on many downstream multi-modal tasks.  
From the experimental results shown in \cref{tab:backbones}, we observed that, when combined with our approach, the performance of all backbones has been greatly improved. In particular, when combined with LXMERT, our method achieved a new state-of-the-art overall accuracy, 71.33\% on the VQA-CPv2 dataset. The improvements are remarkable and consistent, which further demonstrates the effectiveness of our method.

\begin{table}[t]
  \setlength\tabcolsep{3pt}
	\centering
	\setlength{\tabcolsep}{1.4mm}{
    \caption{Acc.(\%) results on VQA-CPv2 with respect to different backbones, including SAN, UpDn and LXMERT. $^{\dagger}$ denotes our reproduced results using the open-sourced codes. The best scores are \textbf{bold}. Note that the LXMERT was fine-tuned for 8 epochs.}
    \label{tab:backbones}
    \begin{tabular}{lccccc}
        \toprule
        Model & All   & Yes/No & Num.  & Other & Gap$\Delta\uparrow$ \\
        \midrule
        SAN$^{\dagger}$   & 40.81 & 41.64 & 13    & 48.01 & \multirow{2}[2]{*}{\textcolor[rgb]{ 0,  .69,  .314}{\textbf{+14.61}}} \\
        SAN+ours  & \textbf{55.42} & \textbf{83.83} & \textbf{24.07} & \textbf{49.14} &  \\
        \midrule
        UpDn$^{\dagger}$  & 41.48 & 43.22 & 13.41 & 48.27 & \multirow{2}[2]{*}{\textcolor[rgb]{ 0,  .69,  .314}{\textbf{+21.38}}} \\
        UpDn+ours  & \textbf{62.86} & \textbf{89.18} & \textbf{55.86} & \textbf{50.98} &  \\
        \midrule
        LXMERT & 46.23 & 42.84 & 18.91 & 55.51 & \multirow{2}[2]{*}{\textcolor[rgb]{ 0,  .69,  .314}{\textbf{+25.10}}} \\
        LXMERT+ours & \textbf{71.33} & \textbf{80.82} & \textbf{59.45} & \textbf{69.62} &  \\
        \bottomrule
    \end{tabular}%
  }
\end{table}

\begin{table}[t]
	\setlength\tabcolsep{3pt}
	\centering
	\setlength{\tabcolsep}{2.7mm}{
		\caption{Effects of the  $\mathcal{L}_{apt}$ and the $\mathcal{L}_{kd}$ components. The results Acc.(\%) are on VQA-CPv2 test split, and the backbone model is UpDn.}
		\label{tab:abli}
		\begin{tabular}{cccc}
			\toprule
			& $\mathcal{L}_{apt}$ &$ \mathcal{L}_{kd}$  & All\\
			\midrule
			1	&  &   &41.48 \\
			2	& \checkmark  &   &42.40 \\
			3	&  &\checkmark  & 62.43\\
			4   &\checkmark  &\checkmark  & \bf62.86  \\
			\bottomrule
		\end{tabular}
		}
\end{table}

\begin{figure}[t]
	\centering
	\includegraphics[width =.42\textwidth,height=0.21\textwidth]{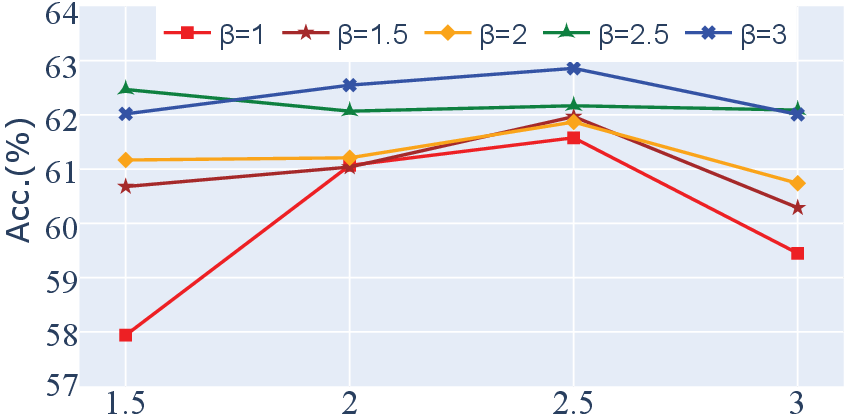}
	\caption{Effect of the hyperparameters scalar weight $\beta$ and temperature $\tau$. The $x$-axis measures the temperature and the $y$-axis quantifies the accuracy.}  
	\label{Fig3}
\end{figure}

\noindent{\textbf{Ablation Study.}}
The effectiveness of $\mathcal{L}_{apt}$ and $ \mathcal{L}_{kd}$ in our proposed method is shown in \cref{tab:abli}. From these results, we can see that when using $\mathcal{L}_{apt}$, the accuracy improvement surpasses the baseline UpDn with 0.92\%, which means the proposed re-weighted method can help to reduce \textit{distribution bias}. While the $ \mathcal{L}_{kd}$ is used for evaluation, the absolute improvement arrives at +20.95\% compared with UpDn, a relative improvement of +0.52\% over the teacher model. The results can reflect that the language bias in VQA task can be alleviated via the benefits of soft labels of the teacher model. Moreover, we can see that when both $\mathcal{L}_{apt}$ and $ \mathcal{L}_{kd}$ used, the best overall accuracy holds.
Besides, we evaluate the effect of different combinations between hyperparameters, scalar weight $\beta$ and temperature $\tau$ on VQA-CPv2. The results are depicted in \cref{Fig3}, we can see that the accuracy of different $\beta$ exhibits a tendency of increasing firstly and then diminishing as the temperature, $\tau$ rises. When $\beta=3$ and $\tau=2.5$, our method builds the best overall accuracy. 

\begin{table}[!ht]
	\setlength\tabcolsep{3pt}
	\centering
	\setlength{\tabcolsep}{2.5mm}{
		\caption{ Acc.(\%) results comparisons with respect to different answer types on VQAv2 val split. The best scores are \textbf{bold}.}
		\label{tab:vqav2}
		\begin{tabular}{lcccc}
			\toprule
			Model & All & Yes/No &Num. & Other\\
			\midrule    
			SAN\cite{yang2016stacked}  &52.41 &70.06 &39.28 &47.84\\ 
			UpDn\cite{anderson2018bottomup} &63.48 &81.18 &42.14 &55.66 \\ 
			GVQA\cite{agrawal2018dont} &48.24 &72.03 &31.17 &34.65\\ 
   
			\midrule 
			SCR\cite{wu2019self-critical} &62.30 &77.40 &40.90 &56.50\\
			HINT\cite{selvaraju2019taking} &63.38 &81.18 &42.14 &55.56\\
			SSL\cite{zhu2020overcoming} &63.73  & - & - & - \\
			CSS\cite{chen2020counterfactual} &59.91 &73.25 &39.77 &55.11\\
			ESR\cite{robik2020a}  &48.90 &69.80 &11.30 &47.80 \\
			MUTANT\cite{gokhale2020mutant} &62.56 &82.07 &42.52 &53.28 \\    
            D-VQA\cite{wen2021debiased} &64.96 &82.18 &44.05 &57.54 \\
           
            SQT\cite{huai2023sqt}  &65.20  & 82.34  & 44.98  & 57.52 \\
			\midrule 
            RUBi\cite{cadene2019rubi} &44.23 &67.05 & 17.48 & 39.61\\
            LMH\cite{clark2019dont}  &56.35 &65.06 &37.63 &54.69\\  
            DLR\cite{jing2020overcoming} &48.87 &70.99 &18.72 &45.57\\ 
			GGE\cite{han2021greedy} &59.11 &73.27 &39.99 &54.39 \\
            GGD\cite{han2023general} &62.15 &79.25 &42.43 &54.66 \\
			CF-VQA\cite{niu2021counterfactual} &63.54 &82.51 &43.96 &54.30\\
			Unshuffling\cite{teney2021unshuffling} &61.08 &78.32 &42.16 &52.71\\
			AdaVQA\cite{guo2021adavqa} & 54.67 &72.47  &\textbf{53.81} & 45.58\\

            IntroD\cite{niu2021introspective}  &63.40     & 82.48     & 46.60     & 54.05 \\
            RMLVQA\cite{basu2023rmlvqa}  &59.99     & 76.68     & 37.54     & 53.26 \\
            MDDC\cite{li2023multi}  &63.33     & 81.64     & 42.56     & 54.88 \\
			Ours &\bf65.54	&\bf82.65	&45.16	&\bf57.91 \\
			\bottomrule
	\end{tabular}}
\end{table}

\subsubsection{Performance on VQAv2}
Also, we evaluate the proposed method on the independent-identical-distribution (IID) scenario, \ie VQAv2, the experimental results are shown in \cref{tab:vqav2}, our method performs best across all question types except Num. type.
The results in both IID (VQAv2) and OOD (VQA-CPv2) scenarios show that our proposed method can boost the generalization of VQA models under different data distribution scenarios.  

\section{Conclusion}
In this paper, we propose a novel approach to alleviate the language priors issue in the VQA task. Specifically, we introduce two learning strategies to mitigate the language prior-dependency. By reducing the models' focus on common samples while increasing their attention to rare samples, the models can learn more comprehensive and generalizable knowledge. 
In addition, our method is generic which can be easily combined with the existing VQA models to boost their abilities. Extensive experiments on VQA-CPv2 and VQAv2 validate the effectiveness of our method.
However,  we solely investigated the case where the debiased model acted as the teacher model. 
In the future, we will probe solutions that incorporate both biased and debiased models simultaneously as teacher models to further enhance the generalization.

\section{Acknowledgements}
This work was supported in part by the National Natural Science Foundation of China under Grant No. 62276110, No. 62172039 and in part by the fund of Joint Laboratory of HUST and Pingan Property \& Casualty Research (HPL). The authors would also like to thank the anonymous reviewers for their comments on improving the quality of this paper.

\bibliographystyle{IEEEbib}
\bibliography{ref}

\end{document}